\documentclass[letterpaper, 10 pt, conference]{ieeeconf}  

\IEEEoverridecommandlockouts                              

\usepackage{amsmath} 
\usepackage{amssymb}  

\usepackage{graphicx,color}
\usepackage[caption=false,font=scriptsize,format=hang]{subfig}
\usepackage{booktabs}
\usepackage{multirow}
\usepackage{url}
\usepackage[sort,nocompress]{cite}
\usepackage{tablefootnote}

\usepackage{xspace}
\makeatletter
\DeclareRobustCommand\onedot{\futurelet\@let@token\@onedot}
\def\@onedot{\ifx\@let@token.\else.\null\fi\xspace}

\def\eg{\emph{e.g}\onedot} 
\def\ie{\emph{i.e}\onedot}

\def\etal{\emph{et al}\onedot}
\makeatother

\makeatletter
\let\NAT@parse\undefined
\makeatother
\usepackage{hyperref}

\graphicspath{{./imgs/}}

\title{\LARGE \bf
Prompter: Utilizing Large Language Model Prompting for a Data Efficient Embodied Instruction Following
}

\author{Yuki Inoue$^{1}$ and Hiroki Ohashi$^{1}$
\thanks{$^{1}$Intelligent Vision Research Dept., Hitachi Ltd., Tokyo, Japan
        {\tt\small \{yuki.inoue.wh, hiroki.ohashi.uo\}@hitachi.com}.}%
}

\begin{document}

\maketitle
\thispagestyle{empty}
\pagestyle{empty}

\begin{abstract}

Embodied Instruction Following (EIF) studies how autonomous mobile manipulation robots should be controlled to accomplish long-horizon tasks described by natural language instructions.
While much research on EIF is conducted in simulators, the ultimate goal of the field is to deploy the agents in real life.
This is one of the reasons why recent methods have moved away from training models end-to-end and take modular approaches, which do not need the costly expert operation data.
However, as it is still in the early days of importing modular ideas to EIF, a search for modules effective in the EIF task is still far from a conclusion.
In this paper, we propose to extend the modular design using knowledge obtained from two external sources.
First, we show that embedding the physical constraints of the deployed robots into the module design is highly effective. Our design also allows the same modular system to work across robots of different configurations with minimal modifications.
Second, we show that the landmark-based object search, previously implemented by a trained model requiring a dedicated set of data, can be replaced by an implementation that prompts pretrained large language models for landmark-object relationships, eliminating the need for collecting dedicated training data.
Our proposed Prompter achieves 41.53\% and 45.32\% on the ALFRED benchmark with high-level instructions only and step-by-step instructions, respectively, significantly outperforming the previous state of the art by 5.46\% and 9.91\%.
The code will be released upon acceptance.

\end{abstract}

\section{Introduction} \label{sec:intro}

The development of autonomous agents that can follow language instructions has long been one of mankind’s dreams.
Embodied Instruction Following (EIF) studies just that- mobile agents are given natural language instructions to perform long-horizon tasks involving both navigation and interactions with the environment.
Unlike similar tasks in the field of embodied intelligence, EIF requires agents to not only navigate unseen environments but also interact with the environments. This makes variations in tasks, language instructions, and the action space that agents must handle significantly more complex. As a result, EIF is considered one of the most human-like and challenging tasks.

Most early attempts in EIF trained end-to-end models using imitation learning \cite{alfred,suglia2021embodied,pashevich2021episodic}, but because imitation learning is relatively data-hungry and requires expert trajectory data, they were eventually replaced by modular approaches which are both sample-efficient and high performing \cite{blukis2022persistent,film,liu9planning,murray2022following,liu2022lebp}.
As the ultimate goal of EIF is to deploy agents in real life, this switch to reduce the data cost and ease the sim-to-real transition is only natural.
Since then, various modules are proposed to improve the performance \cite{liu9planning,murray2022following,liu2022lebp}, but most require extra training data which partially negates the sample efficiency of modular approaches.

In this work, we propose Prompter, which extends previous modular EIF systems with knowledge obtained from external sources.
First source of knowledge is the physical constraints of the agents, such as their dimensions and physical capabilities. One tricky aspect of EIF is that it involves interactions with the environment, so it is critical to incorporate the agent's physical constraints to the modular design.
While this seems like a basic idea, many EIF studies have overlooked this point. We show that a significant performance improvement can be achieved from this point alone.
In addition to the performance gain, as robot's physical constraints are easily obtainable in any real or simulator environment, our modules can be easily customized for agents of different configurations.

Second source of knowledge is the pretrained large language models (LLMs). It has been shown empirically that by using appropriately formatted prompts, common sense information can be retrieved from LLMs \cite{petroni2019language}, and some studies have successfully applied the retrieved information in robotic settings \cite{ahn2022can,huang2022language,majumdar2022zson,gadre2022clip,al2022zero}. The greatest benefit of LLM prompting is that it can be done in a zero-shot manner, requiring no extra data for fine-tuning.
However, unlike previous studies which apply LLM prompting for zero-shot image recognition \cite{majumdar2022zson,gadre2022clip,al2022zero} and instruction understanding \cite{ahn2022can,huang2022language}, we propose to utilize it for landmark-based object search, which is a module that predicts the locations of the unobserved objects from the locations of the observed landmark objects.
Landmark-based search has been shown to be effective in EIF tasks \cite{film,murray2022following,ramakrishnan2022poni}, but the module is implemented by a model trained on a dedicated dataset, which is not data-efficient and also prone to simulator bias.
By implementing the module with LLM prompting, we realize zero-shot prediction.

To summarize, our contributions are three-fold: 
(1) We propose Prompter, a modular approach to the EIF task that incorporates external knowledge sources such as agents' physical constraints.
Although physical constraints information have been used implicitly in previous EIF research, to our knowledge this is the first study that explicitly examines their effect. We also believe that by explicitly listing how each constraint is used, it can act as a ``checklist'' that others in the field can adopt in their task setting.
(2) We also propose to utilize LLMs as a source of external knowledge, to implement landmark-based object search in a zero-shot manner. To our knowledge, this is the first attempt to utilize LLMs for extracting inter-object relationships, and we show that it has the potential to match the performance of the models trained on dedicated datasets.
(3) We evaluate Prompter on ALFRED \cite{alfred}, one of the most challenging EIF benchmarks, and show that it outperforms the previous state of the art by significant margins.

\section{Related Work}

\subsection{Embodied Intelligence}

There are many tasks in the field of embodied intelligence, including Vision Language Navigation (VLN) in which agents are tasked to navigate through unknown environments using language input and visual feedback \cite{anderson2018vision,fried2018speaker,zhu2020vision}, and Embodied Question Answering (EQA) in which agents must answer questions about their surroundings \cite{das2018embodied,bisk2020experience}.
Unlike VLN and EQA, EIF requires an agent to execute multiple subtasks in one run, many of which involve interactions with the environment, making EIF a challenging task.

Our proposed Prompter is evaluated on the ALFRED benchmark. ALFRED is one of the only EIF benchmarks that requires the deployed agents to handle variety of task types, interaction actions, and object state changes (\eg temperature, cleanliness), making it one of the most realistic EIF benchmarks. Furthermore, it boasts one of the most active communities with a competitive public leaderboard\footnote{\url{https://leaderboard.allenai.org/alfred}}.
Early attempts on ALFRED trained single end-to-end models. For example, Shridhar \etal trained a Seq2Seq model with an attention mechanism \cite{alfred}, while others trained variations of transformers \cite{suglia2021embodied,pashevich2021episodic}.
Recently, modular approaches have been explored, with most equipped with front-end vision and language modules that process raw inputs and a back-end decision-making module that integrates the processed information. In many implementations, the visual information is organized in the form of semantic map, which summarizes the locations of the observed object instances as a top-down view of the scene \cite{film,liu9planning,murray2022following,blukis2022persistent}.
However, as this map is built with partial and erroneous observations of the environment, the decision-making module must be carefully designed, giving rise to the differences among the methods. Blukis \etal implement it as a trainable module \cite{blukis2022persistent}, Min \etal's FILM and Murray \etal add the semantic search module to complement the partial information \cite{film,murray2022following}, and Liu \etal invest first half of each run for information gathering, so that the map is as complete as possible \cite{liu9planning}.


\subsection{Large Language Models (LLMs)}  \label{ssec:llm_works}

In this paper, we propose a module based on prompting off-the-shelf LLMs.
Prompting is based on the idea that because LLMs are trained to complete sentences, desired information can be retrieved by providing an LLM with partially-completed sentences. For example, if one wants to know the capital of Japan, an incomplete sentence ``Capital of Japan is'' is fed to an LLM. The query result is obtained by checking how LLM ends the sentence.

Because LLMs are trained on huge corpus that are subject agnostic, LLMs are empirically shown to be effective source of general knowledge without any domain specific fine-tuning \cite{petroni2019language}.
Due to the high generality and a relatively high performance, pretrained large language models are starting to be utilized in the field of embodied intelligence.
For example, few studies use LLMs to decompose the high-level language instructions into low-level step-by-step instructions \cite{ahn2022can,huang2022language}.
Others use models such as CLIP \cite{radford2021learning} that can map vision and language information to the same embedding space to realize zero-shot object detection and instance segmentation \cite{majumdar2022zson,gadre2022clip,al2022zero}.

How we utilize LLMs differs from the previous methods in that we use LLMs to quantify how likely it is for a pair of objects to collocate, and use the information to expedite the object search process.
In previous studies, this is achieved by training a dedicated model on a separate dataset \cite{film,murray2022following}, but because LLM prompting is zero-shot, our method significantly increases the data-efficiency of the whole system.


\section{Task Description} \label{sec:task_description}

\begin{figure}[!tb]
	\centering
    \includegraphics[trim={0mm 25mm 0mm 0mm},clip,width=1\linewidth]{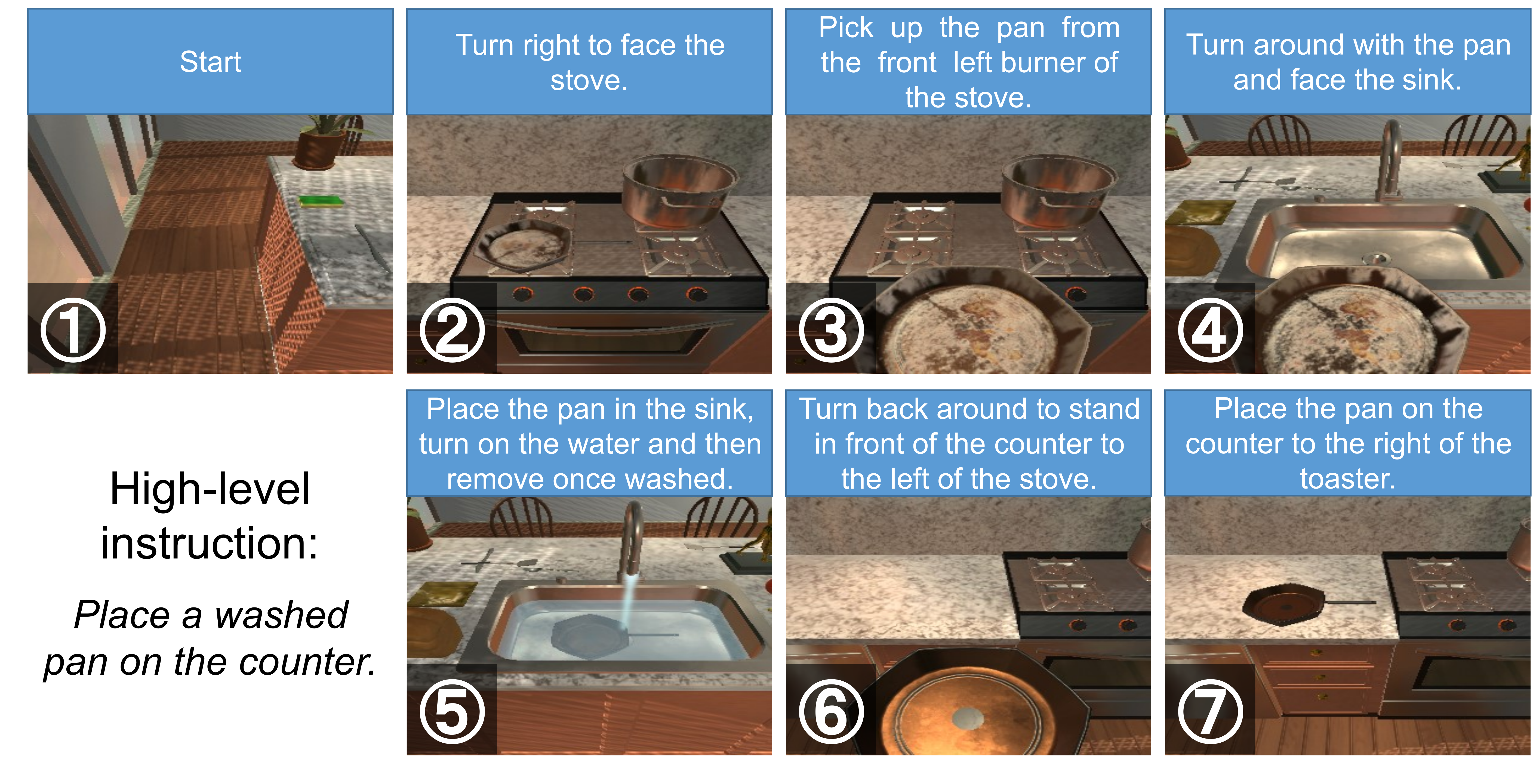}
	\caption{A sample episode in ALFRED. Texts in the blue boxes correspond to the low-level instructions.}
	\label{fig:alfred_example}
	\vspace{-3mm}
\end{figure}

All experiments are conducted on the ALFRED benchmark \cite{alfred}, which is a task based on the iTHOR simulartor \cite{ai2thor} where a mobile manipulation robot must complete long-horizon tasks described by natural language instructions. 
A deployed agent starts an \textit{episode} with no information of the surrounding environment. Only the language instructions that describe the task is available at the beginning of each episode.
There are two types of language instructions in ALFRED. High-level instructions provide the overview of a task, and low-level instructions describe each step.
At each time step, the agent can perform a navigation action \{\texttt{RotateRight}, \texttt{RotateLeft}, \texttt{MoveAhead}, \texttt{LookUp}, \texttt{LookDown}\} or an interaction action \{\texttt{PickupObject}, \texttt{PutObject}, \texttt{OpenObject}, \texttt{CloseObject}, \texttt{ToggleObjectOn}, \texttt{ToggleObjectOff}, \texttt{SliceObject}\}.
Interaction actions must be accompanied by an image mask that specifies the target object.
An episode ends when the task is completed, 1000 steps are taken, or 10 bad interactions are made (\eg collisions, bad target selection).
\ref{fig:alfred_example} shows a sample episode.

\section{Proposed Method} \label{sec:design}

\begin{figure*}[!tb]
	\centering
	\includegraphics[trim={0mm 5mm 0mm 10mm},clip,width=0.95\linewidth]{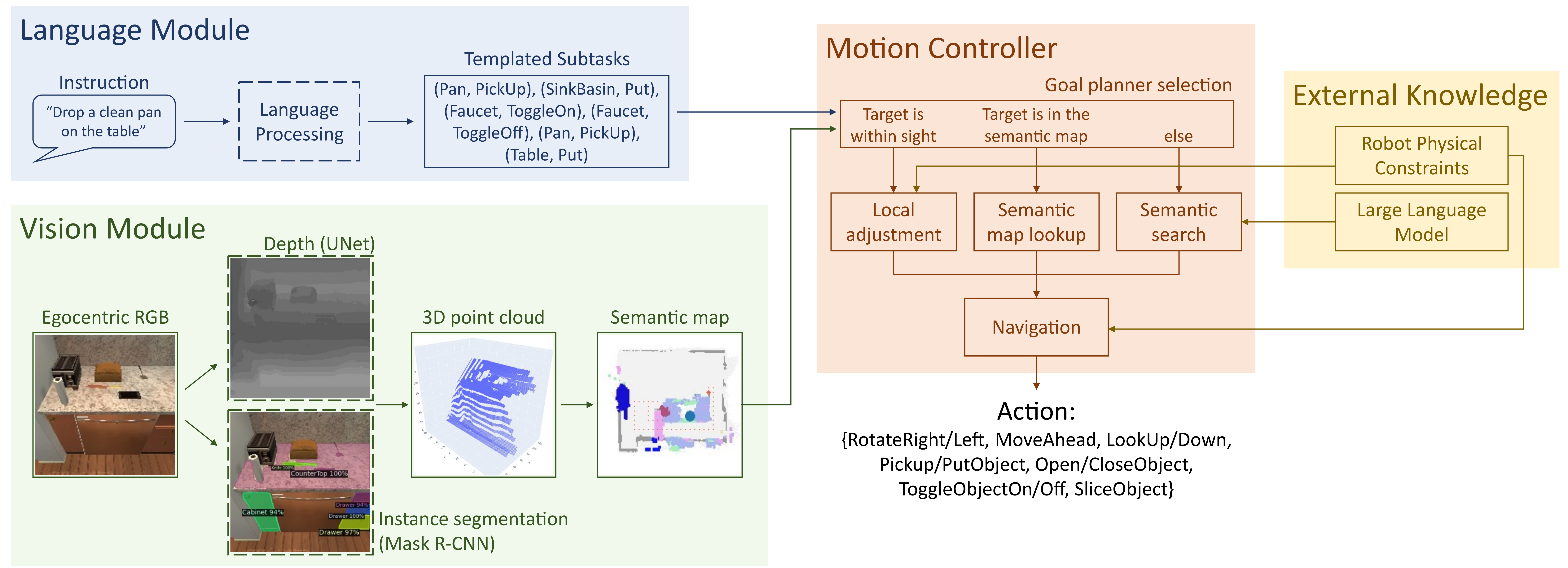} 
	\vspace{-2mm}
	\caption{Method overview. Modules in dotted squares are trained on the ALFRED data. In addition to the processed information from the perception modules, our motion controller utilizes external knowledge such as the physical constraints of the robots and LLM outputs during decision-making.}
	\label{fig:overall}
	\vspace{-2mm}
\end{figure*}

\ref{fig:overall} shows the overview of the proposed method.
First, the free-form language instructions provided at the beginning of an episode and the raw RGB images captured by the agent's camera at each time step are processed by the two perception modules. 
Then, the motion controller integrates the processed information as well as the external information such as the physical constraints of the agent and the LLM to decide the next action.
Some works implement the motion controller as a data-driven module \cite{blukis2022persistent}, but data-driven controllers have a major drawback that they require expert trajectory data for training, which is precisely the reason why modular approaches were selected over end-to-end models to begin with. Furthermore, the controller implicitly learns the physical constraints of the robot, which is undesired as it locks the controller to one specific robot configuration.
Instead of implicitly incorporating the physical constraints into the controller through training, they should be explicitly configurable, as this information is easily obtainable in any environment.
Other works implement the motion controller as rule-based modules like we do in Prompter, but fail to properly incorporate the physical constraints, resulting in sub-optimal performance results.
From the next subsections, we will detail how external knowledge are reflected in the module design.

\subsection{Perception Modules} \label{ssec:perception}

We adopt FILM's \cite{film} design of the perception modules without any modifications.

\subsubsection{Language Module} \label{sssec:language}

The language module decomposes the language instructions into a sequence of object-action pairs, which serve as subtasks that agents follow to complete a task in divide-and-conquer manner. For each object-action pair in the sequence, the agent first searches for the object, and once the object is found, performs the action.
The contents and the order of the object-action pairs are templated for the seven tasks types available in ALFRED. For example, \textit{examine in light} task is always converted to (\texttt{Obj1}, \texttt{PickupObject}), (\texttt{FloorLamp}, \texttt{ToggleObjectOn}), where \texttt{Obj1} is a placeholder replaced depending on the contents of the language instruction.
The task types and the object names are predicted at the beginning of an episode, by several instances of BERTs \cite{devlin2018bert} trained by Min \etal \cite{film}.

\subsubsection{Vision Module} \label{sssec:vision}

The aim of the vision module is to process the egocentric RGB image captured at each time step, and accumulate the information as a 2D top-down map of all observed objects in the scene. This map, named the semantic map, represents the agent's understanding of the surrounding environment.
First, the egocentric image is processed into a depth map via UNet \cite{ronneberger2015u} and an instance segmentation mask via Mask R-CNN \cite{he2017mask}. The depth estimation model trained by Blukis \etal \cite{blukis2022persistent} and the instance segmentation model trained by Shridhar \etal are used without any modifications \cite{shridhar2020alfworld}.
The two estimations are integrated as a semantic 3D point cloud and converted to an allocentric 2D semantic map. The resulting semantic map has the dimension $(C + 1) \times M \times M$, where $C$ is the number of object categories plus one for the obstacle information and $M \times M$ corresponds to the floor space, each cell representing 5 cm $\times$ 5 cm space.

\subsection{Motion Controller} \label{ssec:motion_controller}

The motion planner receives the processed information from the perception modules and plans the next action.
The motion controller first selects the goal planner which plans the next action.
The planner selection process is fully deterministic- (1) if the target object is within sight, the local adjustment module makes final adjustments to prepare the agent for object interaction (Sec. \ref{ssec:local_adjust}), (2) if the target is in the semantic map, the semantic map lookup module guides the agent to the appropriate coordinate, and (3) if the previous two conditions are not met, the semantic search module estimates the location of the target from the observed landmark locations (Sec. \ref{ssec:semantic_search}).
Finally, the navigation module calculates a collision-free path to the coordinate specified by the goal planner (Sec. \ref{ssec:navigation}).



\subsection{Local Adjustment} \label{ssec:local_adjust}

The local adjustment module is selected when the target object is within the agent's sight. It decides if the agent is ready to interact with the target, and if not, relocates it to prepare for the interaction.
Needless to say, an object must be reachable from the agent for it to be interactable.
Two pieces of information are needed to judge if the agent can reach the target object, notably (1) relative position between the target and the agent and (2) given a coordinate, whether the agent have the ability to reach.
The relative position can be calculated by first determining the object location in the camera coordinate system by combining the depth and the instance segmentation predictions, and then converting it to the world coordinate system using the camera position and orientation.
On the other hand, an agent's ability to reach a specific coordinate can be measured prior to the start of the task and supplied as an external knowledge.

Another point of consideration is the distance between the agent and the target object during interactions. In general, it is better to interact with objects at a closer distance, since the vision module tend to be more accurate and object manipulation is also easier. However, this is not be the case for actions that modify the shape of the target, as agents may interfere with the deformed object. For example, when disassembling an object, agents should be positioned to avoid disassembled parts from interfering with the agent. For the cases in which such information is available before the task starts, it should be supplied as external knowledge to appropriately position the agent.

To summarize, this module uses three pieces of external knowledge related to physical constraints- camera pose, agent's reachability and action consequences.
In ALFRED, an agent has a camera at 1.6 m above the ground, and its camera orientation can be obtained from the sensor readings. Furthermore, its manipulation arm can reach any objects that are within 1.5 m from the center of the agent. These pieces of information are used to judge whether the agent is ready for object interaction.
Finally, the only action in ALFRED that modifies the object shape is \texttt{OpenObject}. Considering agents' maximum reach distance of 1.5 m, agents are instructed to position 50 cm away from the target when performing \texttt{OpenObject} action.

\subsection{Semantic Search Module} \label{ssec:semantic_search}

As agents spend much time searching for objects, an efficient search is desired. Many works have suggested the effectiveness of \textit{semantic search module}, which predicts the location of unobserved target object based on observed objects \cite{film,murray2022following,ramakrishnan2022poni}. The module is based on the idea that certain object pairs are more likely to collocate, so observed objects can be used as landmarks to guide the search process. For example, when searching for a fork, one should search near a sink and not near a toilet.
In general, large immobile objects act better as landmarks, as they are easier to spot and less likely to be in arbitrary locations. In ALFRED, objects in the \textit{receptacle} category are selected to be landmarks.

Formally, given $p_x(e_l)$, the probability that landmark $l \in L$ exists at some coordinate $x$ in the scene, the semantic search module predicts $p_x(e_t)$, the probability that object $t$ exists at $x$.
Note that in our setup, $p_x(e_l)$ is approximated by the semantic map calculated in the vision module.
In previous studies, $p_x(e_t)$ is directly estimated by a CNN with trainable parameters $\theta_{\mathrm{cnn}}$, \ie $p_x(e_t) \sim f_{\theta_{\mathrm{cnn}}}([p_x(e_l)]_{l \in L})$. While this approach has been shown to be effective, it has three drawbacks.

First, training $\theta_{\mathrm{cnn}}$ requires ground truths for $p_x(e_l)$ and $p_x(e_t)$, which means that the exact coordinate of every object in the scene must be annotated. Such annotation is easy to collect in a simulator, but time-consuming to collect in real life, as a separate data collection process must be administered.
Second, for the model to be robust in various environments, the ground truths $p_x(e_l)$ and $p_x(e_t)$ must accurately approximate the \textit{natural} distribution of the spatial relationships of objects, which can only be captured after examining many rooms in which organic human activities have taken place. The random object placement by a simulator or artificial rooms prepared by a small group of people will likely form distinctly different, biased distributions.
Finally, the list of objects the model can handle cannot be modified after model training. If one wants to know about the location of some object not defined in the dataset, the data collection and annotation process must be re-administered for that object type. 
This is especially a bad match for the recent direction in the community that tries to replace traditional vision modules with zero-shot models, so that it can be made to recognize objects it has never seen before \cite{majumdar2022zson,gadre2022clip,al2022zero}. Therefore, it would be ideal if the semantic search module can also handle new objects without model retraining.

Instead of training a model to calculate $p_x(e_t)$, we propose to utilize LLM prompting.
First, we expand $p_x(e_t)$ as $p_x(e_t) = \sum_{l \in L} p_x(e_t \mid e_l) \cdot p_x(e_l)$ following Bayes' rule.
$p_x(e_t \mid e_l)$ will be referred to as the \textit{collocation probability}, and it measures the probability of finding object $t$ at landmark $l$. We propose to approximate this distribution using LLM.
Given query template $q$, context samples $c$, and the names of the objects and landmarks as $n_t$ and $n_l$, the collocation probability is estimated via LLM as follows: $p_x(e_t \mid e_l) \sim f_{\theta_{\mathrm{llm}}}(n_t, n_l, q, c)$.
Since $\theta_{\mathrm{llm}}$, the LLM's trainable parameter, is pretrained on a large language corpus, our method solves the three problems mentioned earlier: (1) $\theta_{\mathrm{llm}}$ is pretrained, so training data is not needed (2) the fact that LLMs are pretrained with large corpus remedies the data bias problem (3) can easily adapt to new object types by adding new vocabulary to the list of $n_t$ and $n_l$.

In our experiments, the query template $q$ is: ``If I can search from \texttt{[LANDMARKS]}, I might find \texttt{[$n_t$]} \texttt{[PREP]} \texttt{[$n_l$]}.'' where \texttt{[LANDMARKS]} is a placeholder for the list of all landmarks in a random order, \texttt{[$n_t$]} and \texttt{[$n_l$]} are names of the target object and the landmark with appropriate articles, and \texttt{[PREP]} is either \textit{in} or \textit{on} depending on $n_l$.
In addition, since it has been empirically shown that providing contexts before the final prompt is effective, contexts $c$ are prepared using the query template and sample object landmark pairs. To avoid bias, we made sure that none of the objects in the sample pairs exist in the ALFRED environment. The order of the landmark objects as well as the sample pairs are randomized, and we use the average of 20 runs during $f_{\theta_{\mathrm{llm}}}(n_t, n_l, q, c)$ calculation.
In our experiments, GPT-Neo \cite{gptneo} is used as the LLM.

\begin{table*}[!tb]
	\centering
    \footnotesize
    \setlength{\tabcolsep}{3mm}
    \renewcommand{\baselinestretch}{0.8}
	\caption{Results on the test splits of the ALFRED Benchmark. Bolded values are the top scores for that metric.}
	\begin{tabular}{@{}lcccccccc@{}}
		\toprule
		& \multicolumn{4}{c}{\textbf{Test-Seen}}& \multicolumn{4}{c}{\textbf{Test-Unseen}} \\
		\cmidrule(lr){2-5} \cmidrule(l){6-9}
        \textbf{Method} &SR&GC&PLWSR&PLWGC &SR&GC&PLWSR&PLWGC \\
		\midrule
		\multicolumn{9}{@{}l@{}}{\textbf{Low-level step-by-step instructions + High-level goal instructions}} \\
		\midrule
        Seq2Seq \cite{alfred}               & 3.98 & 9.42 & 2.02 & 6.27 & 0.39 & 7.03 & 0.08 & 4.26 \\
        E.T. \cite{pashevich2021episodic}   & 38.42 & 45.44 & \textbf{27.78} & 34.93 & 8.57 & 18.56 & 4.10 & 11.46 \\
        LWIT \cite{nguyen2021look}          & 30.92 & 40.53 & 25.90 & \textbf{36.76} & 9.42 & 20.91 & 5.60 & 16.34 \\
        FILM \cite{film}                    & 28.83 & 39.55 & 11.27 & 15.59 & 27.8 & 38.52 & 11.32 & 15.13 \\
        LGS-RPA \cite{murray2022following}  & 40.05 & 48.66 & 21.28 & 28.97 & 35.41 & 45.24 & 15.68 & 22.76 \\
        \textbf{Prompter} &\textbf{51.17} & \textbf{60.22} & 25.12 & 30.21 & \textbf{45.32} & \textbf{56.57} & \textbf{20.79} & \textbf{25.8} \\
    	\midrule
		\multicolumn{9}{@{}l@{}}{\textbf{High-level goal instructions only}} \\
		\midrule
        HLSM \cite{blukis2022persistent}&25.11 & 35.79 & 6.69 & 11.53 & 16.29 & 27.24 & 4.34 & 8.45 \\
        FILM \cite{film}                &25.77 & 36.15 & 10.39 & 14.17 &24.46 &34.75  &9.67   &13.13 \\
        LGS-RPA \cite{murray2022following}&33.01&41.71  &16.65  &24.49  &27.80  &38.55  &12.92  &20.01 \\
        EPA \cite{liu9planning}         &39.96 & 44.14 & 2.56 & 3.47 & 36.07 & 39.54 & 2.92 & 3.91 \\
        \textbf{Prompter}               &\textbf{47.95} & \textbf{56.98} & \textbf{23.29} & \textbf{28.42} & \textbf{41.53} & \textbf{53.69} & \textbf{18.84} & \textbf{24.2}\\
        \bottomrule
	\end{tabular}
	\label{tbl:test_eval}
	\vspace{-3mm}
\end{table*}

\subsection{Navigation}\label{ssec:navigation}

The navigation module calculates a collision-free path to the coordinate outputted by the goal planners.
Object collisions can be decomposed into two components, namely the location of obstacles in the scene and the agent size.
In our navigation module, the obstacle information is estimated by the semantic map accumulated in the vision module, and the agent size is provided as the external knowledge.
These two pieces of information are integrated by enlarging the obstacle map with the size of the agent, so that the agent can be modeled as a dimensionless object during path calculation.
In ALFRED, the agent shape can be modeled as a cylinder of radius 20 cm, so the obstacle map is enlarged by 20 cm before the shortest path solver is applied.

\section{Experiments} \label{sec:experiment}

The ALFRED benchmark consists of train (21023 episodes), test-seen (1533 episodes), test-unseen (1529 episodes), val-seen (820 episodes), and val-unseen (821 episodes) sets. The seen splits share the same room layouts as the train split, so the unseen split is used to evaluate how well a model generalizes to new environments.

The methods are evaluated on \textit{Success Rate (SR)} which measures whether all subtasks are completed and \textit{Goal Condition Success (GC)} which measures the ratio of subtasks completed in an episode. Furthermore, both metrics are weighted by (path length of the expert trajectory)/(path length taken by the agent) to produce \textit{path length weighted SR (PLWSR)} and \textit{path length weighted GC (PLWGC)}, quantifying how efficiently tasks are completed.





\subsection{Results}

\ref{tbl:test_eval} shows the results on the test sets. Note that the table is separated by the types of the language instructions used by the agents.
Prompter achieves state of the art on most metrics for both language types, including the test-unseen SR used in the official leaderboard ranking\footnote{The leaderboard has two scores under Prompter. The lower score is reported, as we decided that the run with higher score uses domain knowledge that we felt were too specific to ALFRED.}.
Unseen SR of 41.53\% and 45.32\% present absolute gains of 5.46\% and 9.91\% over previous state of the art, displaying the effectiveness of Prompter.
Also note the significant increase in the path length weighted (PLW) metrics between Prompter and EPA, the previous state of the art for the high-level language instructions. This indicates that Prompter is significantly more efficient, which may become critical in real deployment, as older observations are more likely to get invalidated by other actors in the scene, such as when objects are relocated by humans. 

The only metrics in which Prompter underperforms are the PLW metrics for the test-seen split. Since Prompter has better SR and GC, it implies that E.T. and LWIT finish the tasks considerably faster, as SR and PLWSR should in general scale together. We believe this is caused by overfitting. As landmark objects stay at the same location between the seen splits of train and test, it is possible that these models remembered the landmark locations for a faster task completion. The fact that E.T. and LWIT are implemented as end-to-end models which are prone to overfit and their significant drop in performance from seen to unseen split support this hypothesis.

\subsection{Ablation Study}

Since evaluation on the test split is only available on the leaderboard server, all ablation studies are conducted on the val-unseen split, and all methods are evaluated with high-level instructions only.

\subsubsection{Perception Modules}

\ref{tbl:gt_eval} summarizes how Prompter performs with ground truth language parsing (+ gt lang.), ground truth depth and instance segmentations (+ gt vision), and both (+ gt vision, gt lang.). The table shows that while both modules have much room for improvement, the vision module has a larger room for growth, suggesting that the UNet and the Mask R-CNN models used in the vision module may need to be updated so that future studies on ALFRED are not inhibited by inaccurate image processing. The table also shows human performance on the test-unseen split. Though we cannot make a direct comparison between the human performance and the rest of the table since they are evaluated on different splits, we can see that with strong perception modules, the proposed method may start to reach the human performance.

\begin{table}[!tb]
	\centering
    \scriptsize
    \renewcommand{\baselinestretch}{0.8}
	\caption{Ablation results on ground-truth vision and language.}
	\begin{tabular}{@{}lcccc@{}}
		\toprule
		\textbf{Method}& \textbf{SR}&\textbf{GC}&\textbf{PLWSR}&\textbf{PLWGC} \\
		\midrule
		Prompter                    & 50.18 & 60.64 & 18.31 & 19.94 \\
		+ gt lang.                  & 58.83 & 67.74 & 19.62 & 22.00 \\
		+ gt vision                 & 69.79 & 75.54 & 40.34 & 42.17 \\
		+ gt vision, gt lang.       & 80.02 & 84.10 & 45.27 & 46.58 \\
		\midrule
		Human (test-unseen)         & 91.00 & 94.50 & 85.80 & 87.60 \\
        \bottomrule
	\end{tabular}
	\label{tbl:gt_eval}
\end{table}

\subsubsection{Effect of Physical Constraints}

\ref{tbl:physical_ablation} summarizes the effect of utilizing the physical constraints information, as described in Sec. \ref{ssec:local_adjust} and Sec. \ref{ssec:navigation}.
In \textit{Incorrect obstacle enlargement}, obstacles are enlarged by 10 cm instead of 20 cm during collision-free path planning. 10 cm is chosen here since it is the value incorrectly used in one of the previous methods \cite{film}.
In \textit{Uncorrected reach distance}, the depth estimation by the vision module is directly used to quantify the relative location of an agent and an object, instead of correcting it with the camera pose.
In \textit{No interaction offset}, the agent is always instructed to position directly next to the object during interactions and does not account for object deformation.
The table shows that all modifications improve the performance, especially the obstacle enlargement. This can also be seen in the failure mode analysis summarized in \ref{tbl:failure_analysis}, as Prompter is significantly less likely to fail due to object collision compared to previous methods.


\begin{table}[!tb]
	\centering
    \scriptsize
	\caption{Ablation on utilizing physical constraints information.}
	\begin{tabular}{@{}lcccc@{}}
		\toprule
		\textbf{Descriptions} & \textbf{SR}&\textbf{GC}&\textbf{PLWSR}&\textbf{PLWGC} \\
	    \midrule
		Prompter                    & 50.18 & 60.64 & 18.31 & 19.94 \\
		\midrule
		Incorrect obstacle enlargement       & 28.87 & 40.32 & 12.89 & 17.04  \\
	    Uncorrected reach distance  & 47.99 & 58.34 & 16.80 & 19.32  \\
		No interaction offset      & 47.99 & 58.39 & 16.61 & 18.82  \\
        \bottomrule
	\end{tabular}
	\label{tbl:physical_ablation}
\end{table}


\subsubsection{Semantic Search Module}

\ref{tbl:semantic_search_ablation} summarizes the ablation study on the semantic search module.
\textit{Random search} is a popular baseline \cite{film,murray2022following}, and it randomly samples the next location to search. \textit{CNN} is the $f_{\theta_{\mathrm{cnn}}}$ model described in Sec. \ref{ssec:semantic_search}. The implementation by FILM \cite{film} is used, in which the parameter $\theta_{\mathrm{cnn}}$ is trained on the ALFRED data.

The table shows that random search performs the worst across all metrics. It especially has low PLW metrics, confirming that the semantic search module does expedite task completions. CNN and the proposed LLM prompting perform similarly, which is a satisfactory result for LLM prompting, as it already has three benefits outside of accuracy performance over CNN as described in Sec. \ref{ssec:semantic_search}, namely: (1) no training data is needed (2) less prone to data bias (3) easily extendable to new objects.

\begin{table}[!tb]
	\centering
    \scriptsize
    \setlength{\tabcolsep}{1.5mm}
    \renewcommand{\baselinestretch}{0.8}
	\caption{Ablation results on the semantic search module.}
	\begin{tabular}{@{}lcccc@{}}
		\toprule
		\textbf{Method}& \textbf{SR}&\textbf{GC}&\textbf{PLWSR}&\textbf{PLWGC} \\
		\midrule
		Prompter (LLM prompting) & 50.18 & 60.64 & 18.31 & 19.94 \\
    	Random search       & 48.21 & 58.39 & 17.33 & 19.21 \\
		CNN (FILM)        & 50.67 & 60.58 & 18.88 & 20.97 \\
        \bottomrule
	\end{tabular}
	\label{tbl:semantic_search_ablation}
	\vspace{-3mm}
\end{table}

\subsection{Analysis}

\subsubsection{Collocation Probability} \label{sssec:collocation_prob}


The collocation probability extracted from the LLM is further analyzed to examine how well it captures the inter-object relationship.
The collocation probability is evaluated from three perspectives: (1) does it assign high scores for landmark-object pairs permitted in the simulator? (2) does it assign high scores for landmarks generally good for placing objects? (3) when objects are hidden inside of landmarks, can it be used to estimate which landmark is most likely hiding the object?

The first two metrics, Probability Ratio (PR) and Receptacle Ratio (RR), are based on the iTHOR simulator's definition of object-landmark pairs that are allowed to be positioned together. For example, a baseball bat can be placed on a bed, but not in a refrigerator. Though not perfect, this definition roughly captures the collocation relationship we want to extract from the LLMs.

For convenience, let us define \textit{Collocation Mean Operator} ($\mathcal{M}$) which takes the mean of the collocation probability for some set of object-landmark pairs $(t, l) \in \mathcal{A}$:

\begin{equation*}
    \mathcal{M}(\mathcal{A})  = \frac{1}{|\mathcal{A}|} \sum_{(t, l) \in \mathcal{A}} p_x(e_t \mid e_l)
\end{equation*}

Defining the set of all object-landmark pairs allowed in iTHOR as $\mathcal{G}$ and not allowed as $\mathcal{G}^\complement$, PR is defined as follows:

\begin{equation*}
    PR = \frac{ \mathcal{M}(\mathcal{G}) } { \mathcal{M}(\mathcal{G}^\complement) }
\end{equation*}

Higher PR means that the collocation probability $p_x(e_t \mid e_l)$ tends to assign higher scores for object-landmark pairs permitted in the simulator.

One of the desired properties of collocation probabilities is to capture which landmarks are generally good for object search. Such information would provide a good backup plan for the case in which the target object is not found on the initial try.
To quantify this point, the 24 landmark objects in ALFRED are sorted by the number of object-landmark pairs that are allowed in the simulator, and the top 25\% landmarks are selected to form the set $\mathcal{L}$ and $\mathcal{L}^\complement$ otherwise to define RR:


\begin{equation*}
    RR = \frac{ \mathcal{M}(\mathcal{L}) } { \mathcal{M}(\mathcal{L}^\complement) }
\end{equation*}

The two metrics are used to evaluate the extracted collocation probabilities, as summarized in \ref{tbl:llm_ablation}.
Prompter maintains high scores for both PR and RR, especially against \textit{Uniform}, which models the collocation probability as a uniform distribution.
We also ablate against different numbers of context samples. It shows that as the number of contexts is increased, RR improves while PR stays constant.
We hypothesize that without context samples, LLM does not recognize that the prompt is about object-landmark collocation. As a result, it tends to score high for object-landmark pairs that are conceptually similar, such as book and shelf, but is unable to identify landmarks generally good for object search, such as the countertop.

The last metric on the table, \textit{KL Divergence}, measures the potential of the LLM prompting to be useful in future works.
One of the issues with most systems in EIF, including Prompter, is that it is impossible to find objects hidden in container-like landmarks.
The difficulty with hidden objects is that the agent must recognize which object is most likely to contain the hidden object. For example, a basketball should be looked inside a cabinet, not inside a fridge.
We show here that the LLM prompting approach has the potential to be utilized for this purpose.
Of the 24 landmarks, six landmarks that are capable of hiding objects are selected, and from the val-unseen split of ALFRED, we extract the probability distribution of different objects hidden in those six landmarks.
This pseudo-ground truth distribution is compared against the LLM collocation probability via KL divergence to quantify how well the LLM collocation probability estimates the locations of the hidden objects.
\textit{KL Divergence} column in \ref{tbl:llm_ablation} summarizes the result, showing that the LLM prompting is better than a uniform random guess.

	

\begin{table}[!tb]
	\centering
    \scriptsize
    \setlength{\tabcolsep}{1.5mm}
    \renewcommand{\baselinestretch}{0.8}
	\caption{Ablation results on the collocation probability extraction. \textit{Uniform} stands for uniform distribution. $\uparrow$ indicates higher values are better and vice versa.}
	\begin{tabular}{@{}lcccc@{}}
		\toprule
		\multirow{2}{*}{\textbf{Method}}& \textbf{Num.} & \textbf{Probability}&\textbf{Receptacle}&\textbf{KL Divergence} \\
		&\textbf{Contexts}&\textbf{Ratio ($\uparrow$)}&\textbf{Ratio ($\uparrow$)}&\textbf{($\downarrow$)} \\
		\midrule
    	Uniform       & - & 1 & 1 & 1.45  \\
		\midrule
    	\multirow{4}{*}{Prompter}   & 0 & 2.07 & 0.82 & 0.98  \\
    	                            & 1 & 1.84 & 0.85 & 0.92  \\
    	                            & 5 & 1.82 & 1.46 & 1.06  \\
    	                            & 10 (proposed)& 1.91 & 1.76 & 1.05  \\
        \bottomrule
	\end{tabular}
	\label{tbl:llm_ablation}
	\vspace{-3mm}
\end{table}

\subsubsection{Error Modes}

\ref{tbl:failure_analysis} shows the common error modes of Prompter and the previous method, FILM \cite{film}.
The table shows that over half of Prompter's errors correspond to \textit{Goal object not found} or \textit{Language processing error}, which are errors in the perception modules. Prompter is particularly bad at recognizing small objects such as salt shakers, and large objects that are difficult to recognize up close, such as refrigerators and floor lamps.
Two notable differences between FILM and Prompter are \textit{Interaction failures} and \textit{Collisions}. We believe the reason why Prompter is more likely to fail during interaction is because \texttt{OpenObject} followed by \texttt{PutObject}, which is arguably the most difficult interaction, typically occurs at the end of tasks. As Prompter is more likely to reach the end of tasks, more interaction failures occur.
Also, the reason for a sharp drop in the number of collisions is due to the better-informed obstacle enlargement (Sec. \ref{ssec:navigation}).

\begin{table}[!tb]
	\centering
    \scriptsize
    \renewcommand{\baselinestretch}{0.8}
	\caption{Failure mode analysis on the val-unseen split. Units in \%.}
	\begin{tabular}{@{}lcc@{}}
		\toprule
		\textbf{Error mode} & \textbf{FILM} & \textbf{Prompter} \\
		\midrule
        Goal object not found       & 26.07 & 28.01 \\
        Interaction failures        & 8.54  & 16.20 \\
        Collisions                  & 11.00 & 0.24 \\
        Object in a closed confinement & 16.16 & 13.03 \\
        Language processing error   & 24.54 & 32.89 \\
        Others                      & 13.69 & 9.62 \\
        \bottomrule
	\end{tabular}
	\label{tbl:failure_analysis}
	\vspace{-4mm}
\end{table}

\subsection{Limitations and Future Works}

The modular framework in EIF is still far from perfect. In this section, we outline the shortcomings of Prompter as well as the ALFRED task itself and suggest some future directions.

\subsubsection{Templated Language Processing}

One of the difficulties in EIF is converting free-form language inputs into step-by-step instructions that agents can follow. As explained in Sec. \ref{sssec:language}, Prompter adopts the templated approach proposed by FILM \cite{film} for this purpose.
While this template approach is effective, it gives rise to situations impossible for the agents to handle.
One major problem is that after the sequence of object-action pairs is fixed at the beginning of each episode, there is no mechanism to update it during the episode.
As a result, it cannot handle the situation in which objects are hidden inside a container, because the \texttt{OpenObject} action needed to discover the hidden object cannot be inserted in the sequence.
Another problem is that the object-action pairs can only store object names. It strips down any other object attributes specified in the original instruction, such as the color and the location of the object.
Therefore, if the scene has multiple instances of the same object type, random instances are selected to complete the tasks, which could be problematic in a real deployment. Unfortunately, most tasks in ALFRED make no distinctions on the specific instance of the objects, leaving this misbehaviour unpunished.
For these reasons, future works should avoid templated language processing and develop language modules that can match the expressiveness of free-form instructions and also capable of modifying its policies mid-episode to reflect the situation.

\subsubsection{Instability of LLM Prompting}

One of the major drawbacks of any prompting-based method is that LLMs are sensitive to prompt templates.
This has been reported in many studies and is an active field of research \cite{petroni2019language,ahn2022can}.
In Prompter, the following features must be handcrafted: (1) rephrasing the names of the objects and landmarks to more natural ones (2) assigning prepositions to landmarks (\eg countertop: on, fridge: in) (3) selecting which objects will be used as landmarks\footnote{Landmark objects are predefined for ALFRED, but that may not be the case in other tasks.} (4) the query template and the sample contexts.
We do remark that while the four points mentioned above may sound a lot, it is still significantly cheaper than having to prepare a dedicated set of data for training the semantic search module, as previously done.

\subsubsection{Size of the iTHOR Rooms}

\begin{figure}[!tb]
    \centering



    \includegraphics[height=28mm]{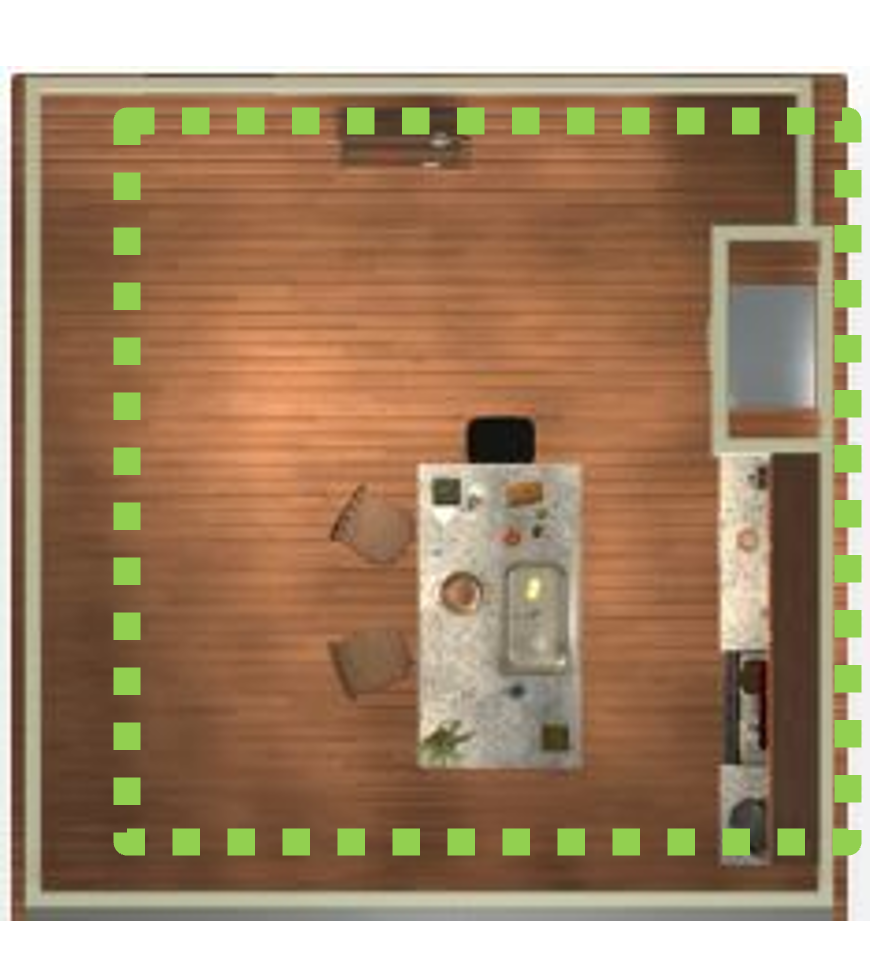}
    \includegraphics[height=28mm]{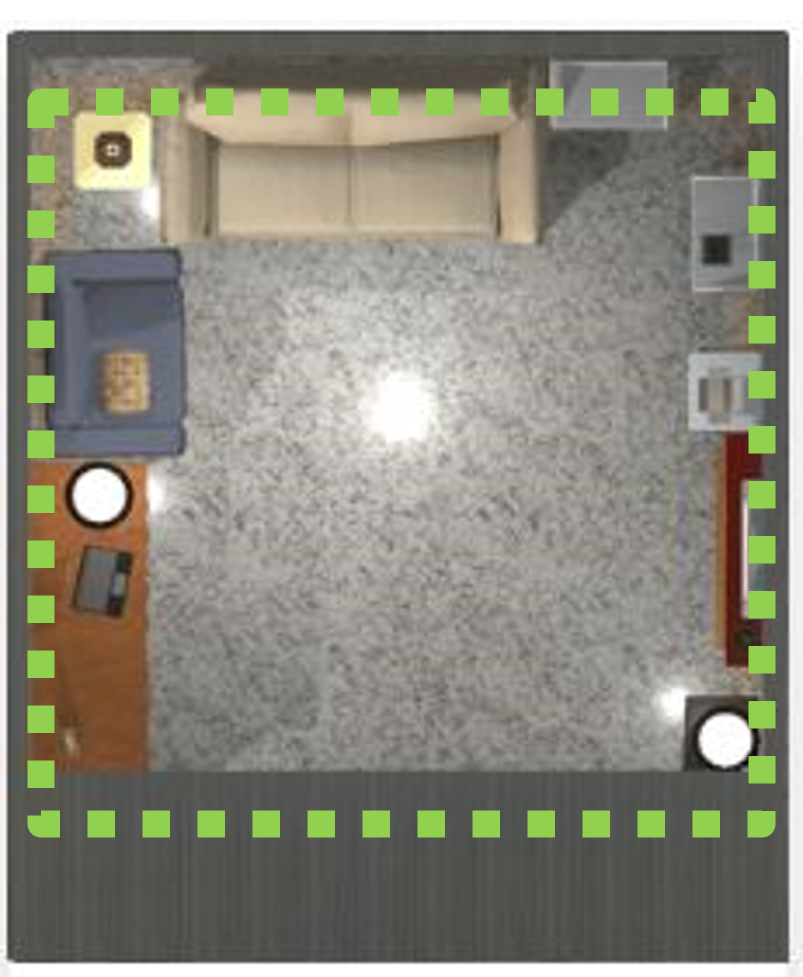}
    \includegraphics[height=28mm]{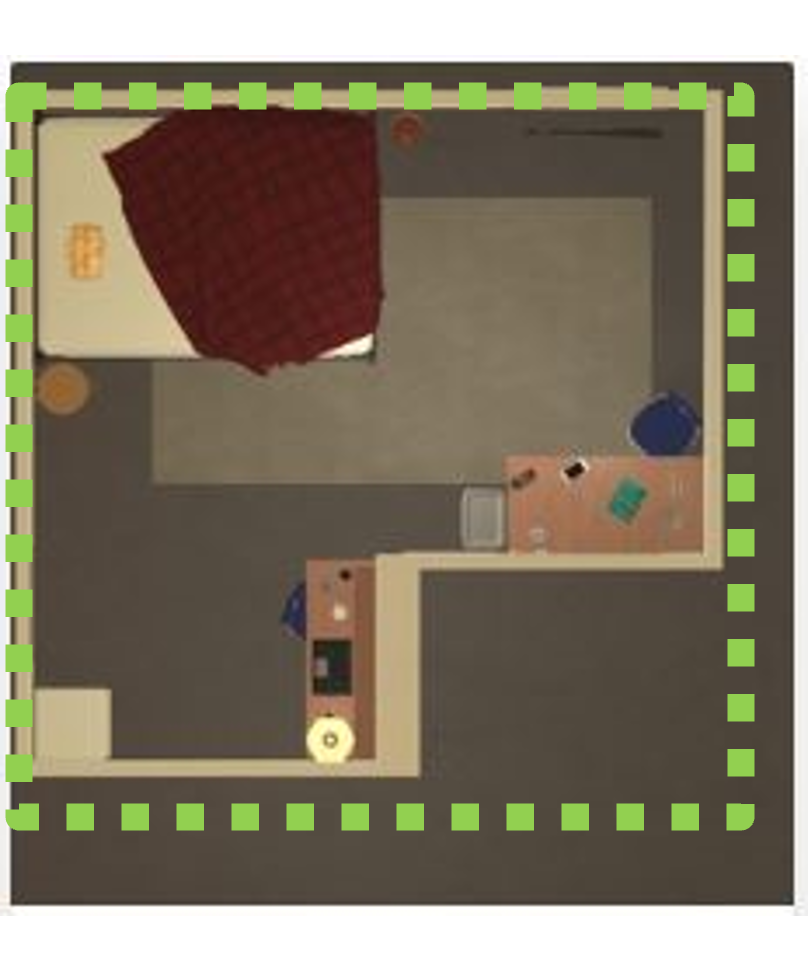}
    \includegraphics[height=28mm]{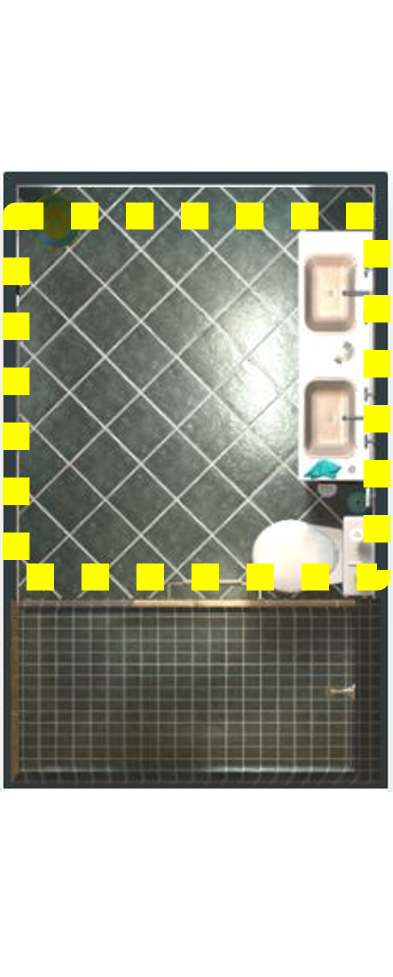}



	\vspace{-3mm}

	\caption{Top-down view of four rooms from the val-unseen split of ALFRED. The images are scaled according to their actual physical sizes. Green and yellow dotted areas in the figures correspond to 5 m $\times$ 5 m and 2.5 m $\times$ 2.5 m areas, respectively. The smallness of the rooms may present unexpected sim-to-real disparities, as most real environments are much larger.}
	\label{fig:val_unseen_rooms}
	\vspace{-3mm}
\end{figure}

The iTHOR simulator used by ALFRED provides a variety of scenes, making it an effective testbed for the EIF tasks. However, iTHOR can only instantiate one room at a time, so all episodes in ALFRED take place in a single room, often only half-filled with objects. As an example, \ref{fig:val_unseen_rooms} shows top-down views of the rooms from the val-unseen split of ALFRED, with green and yellow dotted areas corresponding to 5 m $\times$ 5 m and 2.5 m $\times$ 2.5 m spaces, respectively. As the figure shows, most objects in the rooms can be contained in the dotted space, meaning that the regions that agents must search is quite confined.
This significantly limits the opportunity to challenge agents with more complex situations, and it also indicates that the methods tuned in ALFRED may face difficulties when deployed in reality, as most real environments are much larger. For example, there is a good chance that we are not making accurate comparisons between different semantic search modules, as scenes in iTHOR are too small to properly punish inefficient methods.
One way to resolve this issue is to utilize the recent developments in the community to expand iTHOR to generate multiple rooms for agents to explore \cite{deitke2022procthor}, and create an ALFRED-like EIF benchmark in those environments.

\section{Conclusion}

We proposed Prompter, which augments the previous modular systems designed for EIF with two sources of external knowledge: (1) physical constraints of the deployed agents (2) LLMs for estimating the location of unobserved objects.
Both modifications require little to no data collection and annotation process, making them friendly to new environments. Furthermore, they are general enough to be integrated into other EIF tasks as well.
Finally, Prompter is evaluated on the ALFRED benchmark and showed that it significantly outperforms the previous state of the art.

\bibliographystyle{unsrt}
\bibliography{root}

\addtolength{\textheight}{-12cm}   

\end{document}